\documentclass{article}

\usepackage[preprint, nonatbib]{neurips_2026}
\usepackage{neurips_2025_mods}  % my mods for neurips_2025.sty

\usepackage{amsthm}
\newtheorem{proposition}{Proposition}
\newtheorem{corollary}{Corollary}
\newtheorem{remark}{Remark}

\newcommand{\mat}[1]{\mathbf{#1}}     % shortcut for matrix
\def\rms{\text{RMS}(\mathbf{a})}      % RMS(a)
\def\f1n{\frac{1}{n}}                 % 1/n
\def\sas{\sum_{i=1}^n a_i^2}           % sum over a_i squared
\def\W*{\mat{W}^\ast}                 % matrix W*
\def\V*{\mat{V}^\ast}                 % matrix V*
\def\mV{\mat{V}}                      % matrix V
\def\a{\mathbf{a}}                    % vector a
\def\cosi{\cos{(\cdot)}}              % cos(.)
\def\sini{\sin{(\cdot)}}              % sin(.)

\title{FlashNorm: Fast Normalization for Transformers}

\author{Nils Graef\thanks{\texttt{info@openmachine.ai}}, \, Filip Makraduli, \, Andrew Wasielewski, \, Matthew Clapp \\
  \href{https://openmachine.ai}{OpenMachine.ai}}

\begin{document} \maketitle

\begin{abstract}
Normalization layers are ubiquitous in large language models (LLMs) yet represent
a compute bottleneck: on hardware with distinct vector and matrix execution
units, the RMS calculation blocks the subsequent matrix multiplication, preventing
parallel execution. We present \textbf{FlashNorm}, an \emph{exact} reformulation of
RMSNorm followed by a linear layer that (i) eliminates the normalization weights
by folding them into the subsequent linear layer, and (ii) defers the scalar
RMS normalization to the output of the matrix multiplication, enabling the two
operations to execute in parallel.
%FlashNorm is mathematically identical to RMSNorm---it introduces
%no approximation and requires no retraining.
The same techniques extend to LayerNorm and Dynamic Tanh (DyT).
And due to RMS scale invariance, a linear layer between two RMSNorms makes the
first one redundant, removing the pre-attention RMSNorm in QKV-normalized models (e.g., Gemma~4).
In MLA models with latent normalization (e.g., DeepSeek-V2, Mistral Small 4), its weights
and full-width scaling are eliminated while a per-token scalar is retained on the
decoupled RoPE-key path.
On an NVIDIA T4 GPU, FlashNorm achieves \textbf{33--35\%}
lower latency on the norm-then-project operation in the compute-bound (prefill)
regime at SmolLM2-135M scale, and \textbf{12--14\%} at Llama-7B scale.
We verify zero-loss weight folding on three models and validate the norm
cancellation on Gemma-4-E2B, Gemma-4-12B, MiniCPM3-4B, and DeepSeek-V2-Lite.
Beyond inference speed, FlashNorm simplifies model implementations by
reducing parameter tensor count.
FlashNorm has since been implemented in fused GEMM and attention kernels with up
to $35\%$ speedup \citep{freeNorm}.
Watch our explainer video \citep{flashNorm-video} and see
\citep{tricks} for code.
\end{abstract}

\begin{figure}[h!] \centering  % the [h!] tries to place the picture right here
  \includegraphics[scale=0.7]{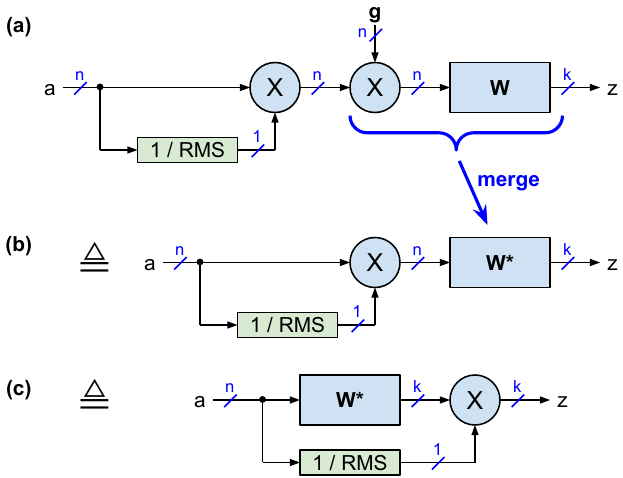}
  \caption{Mathematically identical implementations of RMSNorm followed by a linear layer.
    (a) Standard: normalization weights $\mathbf{g}$ are applied element-wise before $\mathbf{W}$.
    (b) \textbf{Weightless}: $\mathbf{g}$ is folded into $\mathbf{W}^* = \mathrm{diag}(\mathbf{g})\mathbf{W}$.
    (c) \textbf{FlashNorm}: normalization is deferred to the output, enabling matrix multiplication
    ($\mathbf{W}^*$) and RMS calculation to execute in parallel. $\triangleq$ denotes
    mathematical identity.}
\label{fig1} \end{figure}

\section{Introduction}
Normalization is indispensable to the training stability and generalization of
modern LLMs. RMSNorm~\citep{rms}, used by Llama~\citep{LLaMA},
Gemma~\citep{gemma}, Mistral~\citep{mistral}, and
OLMo~2~\citep{olmo2}, normalizes each activation vector by its root mean
square (RMS) before scaling by learned weights. In the standard transformer~\citep{vanilla},
RMSNorm is immediately followed by a linear (projection) layer in both the attention and the
feed-forward sub-network of every block.

This sequential structure hides a hardware inefficiency. On processors with
dedicated vector units (for element-wise ops) and matrix units (for matrix multiplication), the
RMS value must be fully computed \emph{before} the linear layer can begin, because
the normalized activations are its input. The matrix unit therefore sits idle during
the RMS calculation, a bottleneck that compounds across every layer and every inference call.

We address this bottleneck with \textbf{FlashNorm}, which applies two
mathematically equivalent transformations to the RMSNorm-then-project pipeline shown in
Fig.~\ref{fig1}(a):

\begin{enumerate}
  \item \textbf{Weightless normalization.} The per-channel normalization weights
        $\mathbf{g}$ are absorbed into the weight matrix of the following linear layer,
        so they no longer exist as a separate parameter tensor (Fig.~\ref{fig1}(b)).
  \item \textbf{Deferred normalization.} Because scaling by $1/\mathrm{RMS}(\mathbf{a})$
        commutes with matrix multiplication, the scalar normalization can be
        moved to the \emph{output} of the linear layer (Fig.~\ref{fig1}(c)), so the
        matrix multiplication and the RMS calculation can run in parallel on separate
        hardware units.
\end{enumerate}

The name is a nod to FlashAttention \citep{flash-attention}, which similarly
defers the softmax denominator past a matrix multiplication to enable
online (tiled) computation and improve memory efficiency.

FlashNorm requires no retraining: weight folding can be applied post hoc to any pretrained
checkpoint. We also show how the same ideas extend to Layer Normalization~\citep{layerNorm},
DyT~\citep{DyT}, GLU-variant FFNs~\citep{GLU}, and RoPE-based attention~\citep{RoPE}.

\paragraph{Contributions.}
\begin{itemize}
  \item We formally characterize the commutativity of RMSNorm with
        matrix multiplication (Proposition~\ref{prop:commute}) and derive
        the exact conditions under which deferred normalization holds.
  \item We show that RMS scale invariance allows an RMSNorm--linear--RMSNorm
        sequence to drop the first RMSNorm entirely (Proposition~\ref{prop:rms-cancel}),
        eliminating the pre-attention RMSNorm in models with QKV-normalization \citep{qkv-norm}
        such as Gemma~4 \citep{gemma4}. In MLA models with latent normalization
        (e.g., DeepSeek-V2 \citep{deepseek-v2}, Mistral Small 4 \citep{mistral4}),
        the decoupled RoPE-key path bypasses the latent norm, so we give a corrected
        partial transform that retains a per-token scalar on that path only.
  \item We present FlashNorm as a unified framework covering RMSNorm, LayerNorm,
        DyT, GLU-FFNs, and attention with RoPE.
  \item We verify zero-loss weight folding on SmolLM2-135M, Llama-3.2-1B, and
        Llama-3.1-8B across wikitext, MMLU, and HellaSwag (Table~\ref{tab:quality}),
        validate the cancellation on Gemma-4-E2B, Gemma-4-12B, MiniCPM3-4B, and
        DeepSeek-V2-Lite (Table~\ref{tab:cancel}), and report operation-level
        speedups on T4, A100, and H100.
  \item We identify the kernel-engineering work required to extend the fused-kernel
        speedups to compute-bound regimes and provide a reference design.
\end{itemize}

\section{Background and Related Work}

\paragraph{Normalization in transformers.}
Layer Normalization~\citep{layerNorm} applies mean centering followed by
variance normalization with learned scale and bias. RMSNorm~\citep{rms}
drops mean centering and bias, reducing to pure RMS-based scaling; it has become the
dominant normalization choice in recent open LLMs due to its simplicity and
comparable empirical performance. Dynamic Tanh (DyT)~\citep{DyT} is a
recently proposed drop-in replacement that eliminates the normalization computation
entirely, replacing it with a parameterized $\tanh$. FlashNorm is complementary:
it targets the interaction between normalization and the subsequent linear layer.

\paragraph{Kernel fusion and operation reordering.}
Fusing small operations into a single GPU kernel is a standard optimization to
reduce memory bandwidth and kernel-launch overhead. \citet{openelm} report
that modifications to the layer normalization implementation significantly affect
tokens-per-second throughput, attributing this to a lack of kernel fusion. FlashNorm
reduces the number of kernel launches by removing the separate normalization-weight
multiply, and it additionally enables parallelism between the
vector and matrix units that kernel fusion alone cannot provide.
Operation-reordering and epilogue-fusion systems (\citet{opFusion}, CODA~\citep{coda},
Mirage~\citep{mirage}) realize the same deferral at the kernel level while leaving the
stored model unchanged; Appendix~\ref{app:additional-notes} relates them to FlashNorm.

\paragraph{Weight folding and parameter elimination.}
The idea of absorbing one layer's parameters into a neighboring layer appears in
several forms. Folding batch normalization into convolutional weights is standard
practice in quantization-aware training and model deployment~\citep{quantJacob, convBN},
and \verb+torch.compile+ can perform it automatically~\citep{torchCompileFuse}.
OLMo~1 models use weightless LayerNorm, which they call
\emph{non-parametric}~\citep{olmo2}. FlashNorm applies analogous reasoning to
normalization weights of RMSNorm and LayerNorm; see Appendix~\ref{app:additional-notes}
for an extended comparison to \citet{remove-norm2} and OLMo~1.

\paragraph{FlashAttention.}
FlashAttention~\citep{flash-attention} is the closest conceptual predecessor:
it defers the softmax normalization (division by $\sum \exp$) past the multiplication
with the value matrix $\mathbf{V}$, enabling online computation and improving memory
efficiency. FlashNorm applies the same reordering insight to the simpler setting of
RMSNorm and a dense linear layer.

\section{FlashNorm}
We now develop FlashNorm formally. All propositions assume bias-free linear layers
unless stated otherwise; Section~\ref{sec:bias} addresses the biased case.

\subsection{Weightless Normalization}
Let $\mathbf{a} \in \mathbb{R}^n$ be an activation vector. RMSNorm computes
\[
  \mathrm{RMSNorm}(\mathbf{a})_i = \frac{a_i}{\mathrm{RMS}(\mathbf{a})} \cdot g_i,
  \qquad
  \mathrm{RMS}(\mathbf{a}) = \sqrt{\frac{1}{n}\sum_{i=1}^n a_i^2},
\]
where $\mathbf{g} \in \mathbb{R}^n$ are learned normalization weights.
In transformers, RMSNorm is followed by a linear layer $\mathbf{W} \in \mathbb{R}^{n \times k}$ (Fig.~\ref{fig1}(a)):
\[
  \mathbf{z} = \mathrm{RMSNorm}(\mathbf{a})\, \mathbf{W}.
\]

\begin{proposition}[Weightless normalization]\label{prop:weightless}
Let $\mathbf{W}^* \in \mathbb{R}^{n \times k}$ be defined by $W^*_{i,j} = g_i \cdot W_{i,j}$.
Then $\mathbf{z} = \frac{\mathbf{a}}{\mathrm{RMS}(\mathbf{a})}\, \mathbf{W}^*$, i.e.,
the normalization weights $\mathbf{g}$ can be absorbed into $\mathbf{W}$ and need not
be stored or applied separately.
\end{proposition}

\begin{proof}
$z_j = \sum_i \frac{a_i}{\mathrm{RMS}(\mathbf{a})} g_i W_{i,j}
     = \sum_i \frac{a_i}{\mathrm{RMS}(\mathbf{a})} W^*_{i,j} $.
\end{proof}

The proposition holds whether or not the linear layer has an output bias.
In practice, the normalization weight vector $\mathbf{g}$ becomes
a redundant parameter: it can be folded into $\mathbf{W}$ once at load time, which saves
memory, reduces the parameter count, and removes one vector-multiply kernel call
per forward pass.

\subsection{Deferred Normalization}

\begin{proposition}[Commutativity of RMS scaling with matrix multiplication]\label{prop:commute}
Let $\mathbf{W}^*$ be as in Proposition~\ref{prop:weightless} and let the linear
layer be bias-free. Then
\[
  \mathbf{z}
  = \frac{\mathbf{a}}{\mathrm{RMS}(\mathbf{a})}\,\mathbf{W}^*
  = (\mathbf{a}\,\mathbf{W}^*)\cdot \frac{1}{\mathrm{RMS}(\mathbf{a})}.
\]
That is, the scalar normalization $1/\mathrm{RMS}(\mathbf{a})$ can be applied
\emph{after} the matrix multiplication.
\end{proposition}

\begin{proof}
Scalar multiplication commutes with matrix multiplication:
$(\alpha \mathbf{a})\mathbf{W}^* = (\mathbf{a}\mathbf{W}^*) \alpha$ for any scalar
$\alpha$. Setting $\alpha = 1/\mathrm{RMS}(\mathbf{a})$ gives the result.
\end{proof}

\begin{corollary}[Parallel execution]\label{cor:parallel}
Under the conditions of Proposition~\ref{prop:commute}, the computations of
$\mathbf{a}\mathbf{W}^*$ and $\mathrm{RMS}(\mathbf{a})$ are \emph{independent}
and may proceed in parallel. Once both complete, the final result requires only a single
vector-scalar multiply.
\end{corollary}

\begin{remark}[Bias case]
If the linear layer has a bias $\mathbf{c} \in \mathbb{R}^k$, deferred normalization
does not hold directly because
$(\mathbf{a}\mathbf{W}^* + \mathbf{c}) / \mathrm{RMS}(\mathbf{a}) \neq
 \mathbf{a}\mathbf{W}^*/\mathrm{RMS}(\mathbf{a}) + \mathbf{c}$.
In this case, Proposition~\ref{prop:weightless} (weightless normalization) still
applies, but the normalization must precede the bias addition.
\end{remark}

\begin{remark}[Unfolded case]
Proposition~\ref{prop:commute} requires only that $1/\mathrm{RMS}(\mathbf{a})$ be a per-token scalar. Propositions~\ref{prop:weightless} and~\ref{prop:commute} are therefore independent: a framework implementing deferred normalization alone obtains the parallel-execution and fused-kernel speedups (Corollary~\ref{cor:parallel} and Appendix~\ref{app:a100}) for any existing RMSNorm checkpoint, with no weight folding required.
\end{remark}

Figure~\ref{fig1} illustrates the three mathematically equivalent
implementations: (a) standard RMSNorm-then-project, (b) weightless normalization,
and (c) the full FlashNorm with deferred normalization.

\subsection{Cancellation of Pre-Normalization}\label{sec:rms-cancel}

\begin{figure}[h!] \centering
  \includegraphics[scale=0.7]{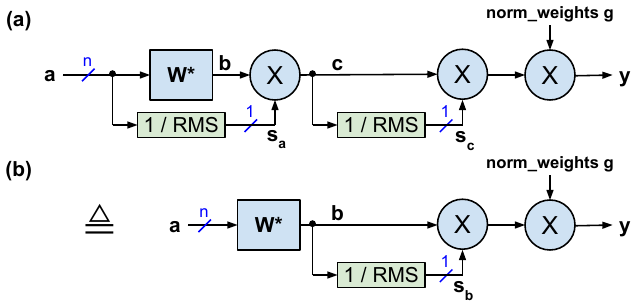}
  \caption{Weightless RMSNorm followed by a linear layer and a second RMSNorm:
    (a) unoptimized version; (b) optimized version where the first RMSNorm is
    eliminated by Proposition~\ref{prop:rms-cancel}.}
\label{fig6} \end{figure}

\begin{proposition}[RMSNorm cancellation]\label{prop:rms-cancel}
For an activation $\mathbf{a}$, a linear layer $\mathbf{W}^*$ (with the normalization weights
already folded in following Proposition~\ref{prop:weightless}), and a second RMSNorm
applied to its output,
\[
  \mathrm{RMSNorm}\!\left(\frac{\mathbf{a}}{\mathrm{RMS}(\mathbf{a})}\,\mathbf{W}^*\right)
  = \mathrm{RMSNorm}\!\left(\mathbf{a}\,\mathbf{W}^*\right),
\]
i.e., the first RMSNorm is redundant and can be eliminated entirely (Fig.~\ref{fig6}).
\end{proposition}
\begin{proof}
Let $\mathbf{b} = \mathbf{a}\mathbf{W}^*$, $s_a = 1/\mathrm{RMS}(\mathbf{a})$, and
$\mathbf{c} = \mathbf{b} \cdot s_a$ (Fig.~\ref{fig6}). By scale invariance of RMS,
$\mathrm{RMS}(\mathbf{c}) = s_a \cdot \mathrm{RMS}(\mathbf{b})$, so the product of the two scaling factors $s_a$
and $\frac{1}{\mathrm{RMS}(\mathbf{c})}$ is $s_a \cdot \frac{1}{\mathrm{RMS}(\mathbf{c})}
= s_a \cdot \frac{1}{s_a \cdot \mathrm{RMS}(\mathbf{b})} = \frac{1}{\mathrm{RMS}(\mathbf{b})}$,
and therefore $\mathrm{RMSNorm}(\mathbf{c}) = \mathrm{RMSNorm}(\mathbf{b})$.
\end{proof}
\paragraph{Eliminating pre-attention RMSNorm with QKV-normalization.}
Some models, such as Gemma~4 \citep{gemma4}, apply QKV-normalization~\citep{qkv-norm} to
queries, keys, \emph{and values}, giving the sequence
pre-attention RMSNorm $\to$ QKV-projection $\to$ QKV-RMSNorm.
By Proposition~\ref{prop:rms-cancel}, the pre-attention RMSNorm cancels and can be
dropped entirely, saving one full RMSNorm per transformer block.

\paragraph{Eligibility condition.}
The pre-attention RMSNorm can be dropped only if \emph{every} consumer of its output
reaches a downstream RMSNorm through a bias-free, positively homogeneous path (a
bias-free linear layer or a slice of one). A consumer that bypasses the downstream norm
would silently change scale, so QK-only-normalized models (e.g., Gemma~3, which has no
value norm) are not eligible.

\paragraph{Pre-attention RMSNorm with MLA latent normalization.}
The sequence for MLA with latent normalization is:
pre-attention RMSNorm $\to$ down-projection into latent space $\to$ latent RMSNorm,
which matches Proposition~\ref{prop:rms-cancel} on the latent path. However, the
decoupled RoPE key is a slice of the same down-projection that bypasses the latent
norm (in DeepSeek-V2 \citep{deepseek-v2}, Mistral Small 4 \citep{mistral4}, and
MiniCPM3 \citep{miniCPM3}), and DeepSeek-V2-Lite additionally projects queries without
a latent norm. Full elimination is therefore not exact for MLA. Instead, we fold
$\mathbf{g}$ into the down-projection (Proposition~\ref{prop:weightless}), drop the
full-width elementwise multiply, and retain the per-token scalar only on the bypassing
slice (32--64 of the projection's output channels). Naively deleting the norm is
catastrophic (Table~\ref{tab:cancel}).

\paragraph{Epsilon caveat.}
Scale invariance, $\mathrm{RMS}(s\cdot\mathbf{a}) = s\cdot\mathrm{RMS}(\mathbf{a})$, does
not hold exactly for the $\varepsilon$-regularized variant
$\mathrm{RMS}_\varepsilon(\mathbf{a}) = \sqrt{\varepsilon + \frac{1}{n}\sum a_i^2}$.
Proposition~\ref{prop:rms-cancel} is therefore exact for unregularized RMS and
$\varepsilon$-approximate otherwise; Appendix~\ref{app:epsilon} gives the exact
per-token discrepancy factor and its measured size, which is negligible unless the
activation energy approaches $\varepsilon$. Propositions~\ref{prop:weightless}
and~\ref{prop:commute} are unaffected, since a per-token scalar commutes with the
matrix multiplication whether or not it contains $\varepsilon$.

\subsection{Support for Normalization Bias}\label{sec:bias}
\begin{figure}[h!] \centering  % the [h!] tries to place the picture right here
  \includegraphics[scale=0.7]{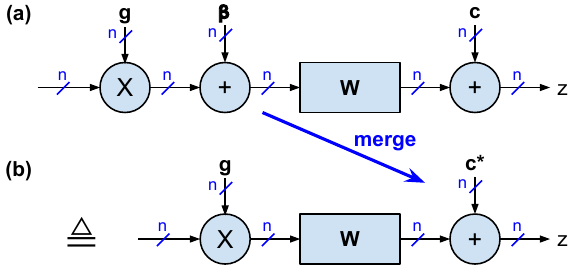}
  \caption{Elimination of bias $\boldsymbol{\beta}$: (a) Before elimination with $\boldsymbol{\beta}$
    between normalization weights $\mathbf{g}$ and linear layer. (b) Optimized version with new
    bias term $\mathbf{c}^* = \mathbf{c} + \boldsymbol{\beta}\mathbf{W}$ at the output.}
\label{figA} \end{figure}

LayerNorm~\citep{layerNorm} and DyT~\citep{DyT} add a
bias $\boldsymbol{\beta}$ after scaling by $\mathbf{g}$. This bias can be moved to
the output of the following linear layer, giving the modified bias
$\mathbf{c}^* = \mathbf{c} + \boldsymbol{\beta}\mathbf{W}$ (Fig.~\ref{figA}).
Once $\boldsymbol{\beta}$ is eliminated, the normalization weights
$\mathbf{g}$ can be folded into the linear layer following Proposition~\ref{prop:weightless}.
Deferred normalization (Proposition~\ref{prop:commute}) does not apply as is
when a bias is present, because the scalar $1/\mathrm{RMS}(\mathbf{a})$ must be applied
before $\mathbf{c}^*$ is added; the matrix multiplication and the RMS calculation
can nevertheless still run in parallel.

%LayerNorm's mean centering can likewise be folded into the preceding linear layer, see Appendix~\ref{app:mean-centering}.

\subsection{Folding Mean Centering into a Preceding Linear Layer}
\begin{figure}[h!] \centering  % the [h!] tries to place the picture right here
  \includegraphics[scale=0.7]{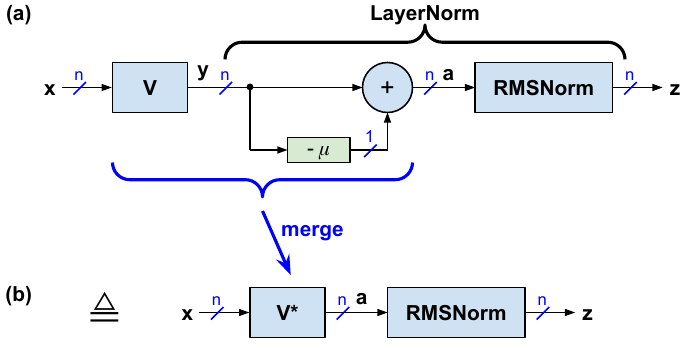}
  \caption{Elimination of mean centering: (a) Original weight matrix $\mV$ followed by mean centering. (b) Optimized version where the mean centering is merged into the modified weight matrix $\V*$.}
\label{figB} \end{figure}

LayerNorm applies mean centering before RMSNorm. If mean centering is preceded by a
linear layer with weights $\mathbf{V} \in \mathbb{R}^{m \times n}$, the centering
can be absorbed into $\mathbf{V}$ without retraining (Fig.~\ref{figB}). The mean $\mu$ is
calculated from the linear layer outputs $y_j$ as
$\mu = \frac{1}{n} \sum_j y_j$, where $y_j = \sum_i x_i V_{i,j}$.
With $s_i = \sum_j V_{i,j}$ denoting the sum of row $i$, we can
calculate $\mu$ directly from the inputs $x_i$ as
\begin{equation*}
  \mu = \frac{1}{n} \sum_{j=1}^n \sum_{i=1}^m x_i V_{i,j} = \frac{1}{n} \sum_{i=1}^m x_i \left[ \sum_{j=1}^n V_{i,j} \right] = \frac{1}{n} \sum_{i=1}^m x_i s_i .
\end{equation*}
The mean-centered outputs are now
\begin{equation*}
  a_j = y_j - \mu = \sum_{i=1}^m x_i V_{i,j} - \frac{1}{n} \sum_{i=1}^m x_i s_i
       = \sum_{i=1}^m x_i \underbrace{\left( V_{i,j} - \frac{s_i}{n} \right)}_{V^*_{i,j}},
\end{equation*}
so the modified weights $V^*_{i,j} = V_{i,j} - s_i/n$ absorb the mean centering
entirely (Fig.~\ref{figB}(b)). This provides a lossless recipe for converting a
LayerNorm model to RMSNorm without retraining; see Appendix~\ref{app:additional-notes}
for the residual-stream completeness condition.

\section{Extensions}

\subsection{FlashNorm for FFNs with GLU Variants}
\begin{figure}[h!] \centering
  \includegraphics[scale=0.7]{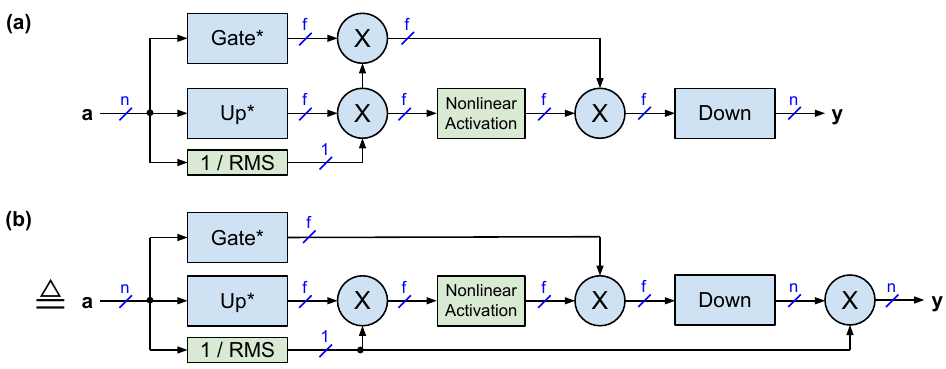}
  \caption{FFN with GLU variant and preceding FlashNorm: (a) unoptimized version; (b) optimized version with fewer scaling multipliers.}
\label{fig3} \end{figure}

For GLU-variant FFNs~\citep{GLU}, applying FlashNorm at the FFN
input requires scaling at both the Gate and Up projection outputs (Fig.~\ref{fig3}(a)).
One of these scaling operations can be deferred to the FFN output, saving $f - n$
multiply operations (Fig.~\ref{fig3}(b)).
Appendices~\ref{app:relu} and~\ref{app:reglu} cover FFNs with ReLU, ReGLU and bilinear GLU.

\subsection{FlashNorm for Attention with RoPE}
\begin{figure}[h!] \centering
  \includegraphics[scale=0.7]{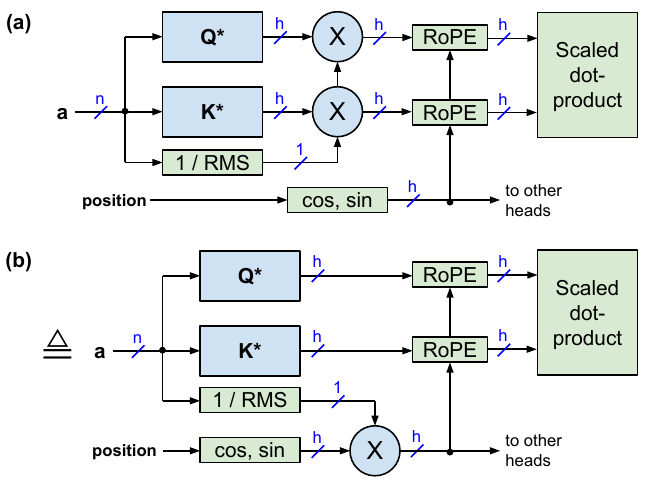}
  \caption{FlashNorm for scaled dot-product attention with RoPE: (a) unoptimized version; (b) optimized version where the normalization is fused with $\cosi$ and $\sini$.}
\label{fig5} \end{figure}

For the query ($\mathbf{Q}^*$) and key ($\mathbf{K}^*$) projections in RoPE-based
attention~\citep{RoPE}, the $1/\mathrm{RMS}(\mathbf{a})$ factor can be
fused with the precomputed RoPE cosine and sine tables, which are shared across
all attention heads (Fig.~\ref{fig5}).
Specifically, replacing $\cos(\cdot)$ with
$\cos(\cdot)/\mathrm{RMS}(\mathbf{a})$ (and similarly for $\sin$) eliminates the
per-head scaling entirely, saving $2hH - h$ multiply operations, where $h$ is the head
dimension and $H$ is the number of heads. The $1/\sqrt{h}$ factor of scaled
dot-product attention can be folded into the same scaling as well.

\subsection{QK-Normalization}\label{sec:qknorm}
\begin{figure}[h!] \centering
  \includegraphics[scale=0.7]{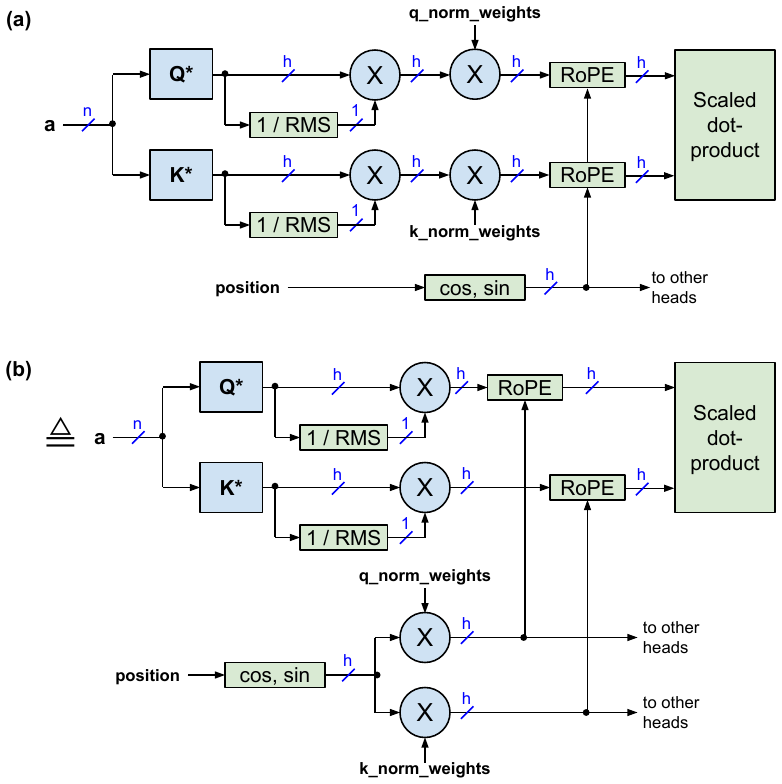}
  \caption{QK-norm with RoPE: (a) unoptimized version; (b) optimized version.}
\label{fig7} \end{figure}

Gemma~3~\citep{gemma3}, OLMo~2~\citep{olmo2}, OpenELM~\citep{openelm} and other LLMs
use QK-normalization~\citep{QKnorm}, which applies RMSNorm to queries and
keys \emph{after} the projection. Unlike QKV-normalization, V is not normalized, so
the pre-attention RMSNorm cannot be fully eliminated, but the $1/\mathrm{RMS}(\mathbf{a})$
scaling of Q and K can still be dropped via Proposition~\ref{prop:rms-cancel}
(Fig.~\ref{fig7}). Two further optimizations are available:

\paragraph{Fusing QK-norm weights with RoPE.}
If QK-norm weights are shared across all heads of a layer (as in Gemma~3 and
OpenELM), they can be fused with the RoPE cosine and sine vectors, which are
already shared across heads (Fig.~\ref{fig7}).
The fused tables are computed once per layer, and
the normalization weights require only a permutation to align with RoPE's
interleaved layout.
% More details: $\mathrm{permute}(\mathbf{g}) = (g_2, g_1, g_4, g_3, \ldots, g_h, g_{h-1})$.

\paragraph{Weight-fusion for NoPE channels.}
Several architectures employ partial RoPE or interleave RoPE and NoPE layers
(e.g., Llama~4~\citep{llama4}). For these models, we can fuse the Q and K normalization
weights for the NoPE channels as
$g_{\mathbf{QK}, i} = g_{\mathbf{Q}, i} \cdot g_{\mathbf{K}, i}$ for $i \in [ \mathrm{NoPE} ]$.

% MORE DETAILS:
%This requires permutation of the normalization weights $\vg$ so that the boxes labeled cos, sin, and RoPE in Fig. \ref{fig7}(b) perform $\vy = \vx \cdot \left( \cosi \cdot \vg \right) + \text{permute}(\vx) \cdot \left( \sini \cdot \text{permuteg}(\vg) \right)$, where $\text{permuteg}(\vg) = (g_2, g_1, g_4, g_3, \dots, g_h, g_{h-1})$. For simplicity, Fig. \ref{fig7}(b) doesn't show the permutation of the normalization weights.

\section{Hardware Analysis}
We analyze the compute bottleneck that FlashNorm eliminates.
Consider a processor with one vector unit
(throughput: $m$ ops/cycle) and one matrix unit (throughput: $m^2$ ops/cycle) where $m$ is the processor width. Then:
\begin{itemize}
    \item Multiplying an $n$-vector with an $n \times n$ matrix requires $n^2$ MAD (multiply-add)
          operations, i.e., $n^2/m^2$ matrix cycles.
    \item Computing  $1/\mathrm{RMS}(\mathbf{a})$ requires $n$ MAD operations (for squaring and adding)
          plus 2 scalar ops (for $\sqrt{n/x}$), i.e., $n/m$ vector cycles if we ignore
          the 2 scalar ops.
    \item Scaling an $n$-vector by a scalar takes $n/m$ vector cycles.
\end{itemize}

\begin{figure}[h!] \centering
  \includegraphics[scale=0.7]{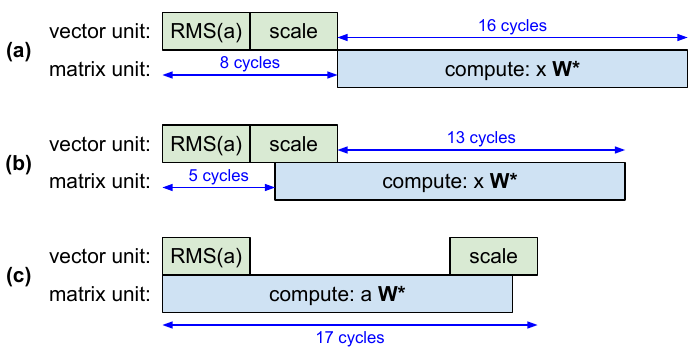}
  \caption{Example timing diagrams.
    (a) Sequential: matrix unit idles 8 cycles.
    (b) Partial overlap with interleaved scaling and vector-matrix multiplication.
    (c) \textbf{FlashNorm}: vector-matrix multiply and RMS run fully in parallel.}
\label{fig8} \end{figure}

Fig.~\ref{fig8} shows timing diagrams for the example $n = 512$ and $m = 128$.
Without FlashNorm (sequential execution), the matrix unit cannot begin until both
the RMS and the scaling are complete, imposing an 8-cycle bubble (Fig.~\ref{fig8}(a)).
The subsequent vector-matrix multiply takes $(512)^2 / (128)^2 = 16$ cycles; the total is 24 cycles.
With row-major ordering, the first rows of $\mathbf{W}^*$ can be processed
while later elements are still being scaled, reducing the bubble
to 5 cycles (21 in total), but the bottleneck persists (Fig.~\ref{fig8}(b)).
With FlashNorm, the vector-matrix multiply ($\mathbf{a}\mathbf{W}^*$) and
the RMS calculation proceed fully in parallel on separate units (Fig.~\ref{fig8}(c)),
and the final scaling can overlap with the matrix unit if $\mat{W}^*$ is processed in column-major order.
The total is 17 cycles, a \textbf{29\% reduction} versus the sequential baseline.

\section{Experiments}\label{sec:experiments}
Refer to \citep{hfFlashNorm, tricks} for Python code that demonstrates the mathematical equivalency of the optimizations presented in this paper. A reproducible Colab notebook for the GPU benchmarks below is available in the \verb+transformer-tricks+ repository\footnote{\url{https://github.com/OpenMachine-ai/transformer-tricks/blob/main/notebooks/flashNorm\_gpu\_benchmark.ipynb}}.

We verified lossless weight folding on SmolLM2-135M \citep{smollm}, Llama-3.2-1B, and Llama-3.1-8B \citep{LLaMA}
using \verb+flashify_repo()+ from the \verb+transformer-tricks+ package \citep{tricks}.
On SmolLM2-135M in HuggingFace Transformers at fp16, the original and weight-folded models produce logits that agree to within fp16 precision (cosine similarity $= 1.0$) and identical greedy generations; at the Llama-3.2-1B and Llama-3.1-8B scales under fp32 inference, greedy generation is likewise identical to the source. Table~\ref{tab:quality} reports source-vs-folded deltas across three model scales and three standard benchmarks (wikitext word perplexity, MMLU 5-shot accuracy, HellaSwag zero-shot accuracy): every delta is within benchmark noise, empirically consistent with the mathematical exactness of Propositions~\ref{prop:weightless} and~\ref{prop:commute}.
Three public compatibility checkpoints are available under \url{https://huggingface.co/open-machine} and load unmodified in stock HuggingFace Transformers and vLLM.

\begin{table}[h!] \centering
\caption{Quality preservation across model scales and standard benchmarks, via \texttt{lm-evaluation-harness} with fp16 inference on an NVIDIA A100 (wikitext at 2048-token context, MMLU at 4096-token context for 5-shot prompts, HellaSwag at 2048-token context). Deltas between source and FlashNorm-folded checkpoints are within benchmark noise at every row, empirically consistent with the mathematical exactness of Propositions~\ref{prop:weightless} and~\ref{prop:commute}.}
\label{tab:quality}
\begin{tabular}{l l r r r}
\hline
Model & Benchmark & Source & FlashNorm & $\Delta$ \\
\hline
SmolLM2-135M & wikitext word\_ppl          & $23.1262$ & $23.1279$ & $+0.001689$ \\
Llama-3.2-1B & wikitext word\_ppl          & $12.8837$ & $12.8842$ & $+0.000497$ \\
Llama-3.1-8B & wikitext word\_ppl          & $8.0668$  & $8.0672$  & $+0.000382$ \\
Llama-3.1-8B & MMLU acc (5-shot)           & $0.6556$  & $0.6562$  & $+0.0006$ \\
Llama-3.1-8B & HellaSwag acc (0-shot)      & $0.6071$  & $0.6072$  & $+0.0001$ \\
\hline
\end{tabular}
\end{table}

\paragraph{Norm cancellation.} We validate Proposition~\ref{prop:rms-cancel} post-hoc on Gemma-4-E2B and Gemma-4-12B (norm removed) and the corrected MLA transform on MiniCPM3-4B and DeepSeek-V2-Lite (Appendix~\ref{app:cancel}, Table~\ref{tab:cancel}). In fp32, cancellation changes wikitext token perplexity by less than $0.001\%$ on both Gemma-4 models, within the rounding envelope of the exact fold-only control, and leaves all 12 greedy continuations unchanged. The corrected MLA transform is likewise within that envelope, whereas naively deleting the norm raises perplexity from 8.8 to 189 (MiniCPM3) and from 7.5 to 618 (DeepSeek-V2-Lite). A matched decode ablation on Llama-3.2-1B (Table~\ref{tab:attribution}) attributes the end-to-end gain of a weightless fused norm kernel to kernel fusion ($+26\%$) rather than to eliding the per-channel multiply ($<1\%$).

To validate the parallel execution described in Fig.~\ref{fig8}, we benchmarked the norm-then-project operation on an NVIDIA T4 GPU (65~TFLOPS FP16, 320~GB/s). We compare sequential execution (RMSNorm $\rightarrow$ GEMM) against FlashNorm with adaptive dispatch: when the GEMM is compute-bound (many tokens), the GEMM runs on tensor cores and $1/\mathrm{RMS}(\mathbf{a})$ runs on vector cores in parallel via separate CUDA streams; when memory-bound (few tokens), FlashNorm falls back to sequential execution to avoid stream overhead. Table~\ref{tab:overlap} shows the results at SmolLM2-135M scale, and Table~\ref{tab:overlap7b} at Llama-7B scale.

\begin{table}[h!] \centering
\caption{Norm-then-project latency on T4 GPU at SmolLM2-135M scale ($n{=}576, k{=}960$). Below the compute-bound threshold, FlashNorm falls back to sequential execution (0\% overhead).}
\label{tab:overlap}
\begin{tabular}{r r r r l}
\hline
Tokens & Sequential & FlashNorm & Latency reduction & Path \\
\hline
1      & 0.374\,ms & 0.374\,ms & $0.0$\% & sequential (fallback) \\
64     & 0.506\,ms & 0.506\,ms & $0.0$\% & sequential (fallback) \\
4096   & 0.704\,ms & 0.468\,ms & $+$33.6\% & parallel (overlap) \\
8192   & 0.929\,ms & 0.599\,ms & $+$35.5\% & parallel (overlap) \\
\hline
\end{tabular}
\end{table}

The parallel path is used only when the GEMM is compute-bound, i.e., when the token count exceeds a hardware-dependent threshold. For memory-bound workloads such as autoregressive decoding, FlashNorm falls back to sequential execution (0\% overhead), because the scheduling overhead of parallel CUDA streams exceeds the time saved on a small GEMM. During autoregressive decoding of \texttt{Llama-3.2-1B-Instruct} in HuggingFace Transformers at bf16, RMSNorm takes 24.6\% of total wall-clock time, an order of magnitude above its FLOP share. Because the cumulative norm cost does not change under weight quantization, the relative impact of FlashNorm should grow wherever low-bit kernels make the linear layers faster; BitsAndBytes on A100, whose low-bit linears are slower than bf16, shows the opposite trend (Appendix~\ref{app:rmsnorm-fraction}).

\section{Limitations and Discussion}\label{sec:limitations-discussion}

\paragraph{Relationship to quantization.}
FlashNorm is orthogonal to weight quantization. In fact, folding normalization
weights into $\mathbf{W}^*$ before quantization can improve quantization quality
by absorbing per-channel scale factors directly into the weight distribution,
a technique already used in practice (e.g., SmoothQuant~\citep{smoothQuant}). We leave a systematic
study of this interaction to future work. Activation dynamic range under fixed-point or
FP8 activations is discussed in Appendix~\ref{app:additional-notes}.

\section{Conclusion}
We presented FlashNorm, a mathematically exact reformulation of RMSNorm-then-project
that (1) eliminates normalization weights by folding them into the subsequent linear
layer, and (2) defers the RMS scaling until after the matrix multiplication,
so that the two can run in parallel.
We further showed (3) that RMS scale invariance allows an RMSNorm--linear--RMSNorm sequence
to drop the first RMSNorm entirely,
removing the pre-attention RMSNorm in models with QKV-normalization (e.g., Gemma~4).
In MLA models with latent normalization (e.g., DeepSeek-V2, Mistral Small 4), a per-token
scalar must be retained on the decoupled RoPE-key path; we validated this corrected
transform on MiniCPM3-4B and DeepSeek-V2-Lite, and the full cancellation on Gemma-4.
GPU benchmarks show a 33--35\% operation-level latency reduction at
the SmolLM2-135M scale and 12--14\% at the Llama-7B scale
in the compute-bound prefill regime, with identical outputs verified on
SmolLM2-135M, Llama-3.2-1B, and Llama-3.1-8B.
In the memory-bound decode regime, stream management overhead
currently negates the speedup, motivating native kernel-level integration as the
primary direction for future work.

Future work includes integration into
vLLM~\citep{vLLM}, SGLang~\citep{sglang}, FlashInfer~\citep{flashInfer},
HuggingFace Transformers~\citep{HFtransformers},
llama.cpp~\citep{llama-cpp}, whisper.cpp~\citep{whisper-cpp},
Ollama~\citep{ollama}, LM Studio~\citep{lmstudio}, and MLX~\citep{mlx}, as well as a systematic
study of the interaction with weight quantization methods.

\section*{Acknowledgments}
We would like to thank Dmitry Belenko for helpful feedback on this work, and Han Guo, co-author of the CODA framework \citep{coda}, for helpful discussions.

\appendix

\section{Additional notes}\label{app:additional-notes}

\paragraph{Scope.} The mathematics of FlashNorm (Propositions~\ref{prop:weightless} and~\ref{prop:commute}) is exact in both directions. Applying FlashNorm post-hoc to a trained model is lossless: the folded model's outputs are identical to the source under fp32 and within benchmark noise under fp16. Training a new model directly with a FlashNorm-shaped normalization-plus-linear block is equally valid: the FlashNorm parameterization (no explicit $\mathbf{g}$, RMS scalar applied after the matmul) represents the same function class as standard RMSNorm followed by a linear layer, so any function reachable under the standard parameterization is reachable under the FlashNorm parameterization. In that setting $\mathbf{g}$ is absent by construction, which also removes the framework-integration work needed to serve post-hoc-transformed checkpoints through existing fused kernels. Experiments in this paper cover the post-hoc application; training-from-scratch studies are left to future work.

The FlashNorm transformation is mathematically exact (Propositions~\ref{prop:weightless} and~\ref{prop:commute}), and under fp32 inference produces identical greedy generation across all three model scales. Under lossy inference precisions (fp16/bf16 in HuggingFace Transformers at larger scales, or vLLM's PagedAttention at any precision), precomputed merged weights can produce a one-token argmax divergence at tight decision points, which downstream greedy decoding then amplifies. The effect is more visible at larger model scales where argmax decisions are tighter. This is a general consequence of precomputing weight-folded tensors for lossy-inference kernels and is not specific to FlashNorm. A native fused \texttt{RMSNorm+QKV} kernel that defers $\mathbf{g}$ to runtime, rather than precomputing it into $\mathbf{W^\ast}$, eliminates the framework dependency and is the primary direction for ongoing work.

\paragraph{Simplification as a primary benefit.}
Beyond raw speed, FlashNorm simplifies model implementations in a manner analogous
to RMSNorm's simplification over LayerNorm, or PaLM's removal of linear-layer
biases~\citep{PaLM}. Fewer parameter tensors mean smaller checkpoint
files, simpler loading logic, and reduced risk of implementation bugs.

\paragraph{Extended comparison to prior weightless / removal lines.} OLMo's non-parametric LayerNorm~\citep{olmo2} is a training-time architectural choice: the model is trained from the start without learnable normalization weights, so there is no transformation of an existing checkpoint. \citet{remove-norm2} target post-hoc LayerNorm removal in GPT-2 XL via an empirically tuned ablation procedure that recovers benchmark quality but is not a closed-form algebraic identity. FlashNorm Proposition~\ref{prop:weightless} differs in three ways relevant to deployment: (i) it is exact in real arithmetic and bit-identical when the fold is performed at the same precision as the runtime, so no quality recovery procedure is required; (ii) it applies to any pretrained RMSNorm checkpoint without modification of training; and (iii) Proposition~\ref{prop:commute} additionally exposes parallel execution of the RMS scalar with the matmul (Corollary~\ref{cor:parallel}), which the prior lines do not address.

\paragraph{Relationship to kernel-level deferral and fusion systems.} \citet{opFusion} defer the Softmax and LayerNorm scaling factors past the subsequent linear layer so that they can be computed concurrently with the matrix multiplication, the same reordering as Proposition~\ref{prop:commute}, presented as a kernel-scheduling technique. CODA~\citep{coda} expresses the memory-bound operations of a transformer block as GEMM epilogues and, for a GEMM--RMSNorm--GEMM chain, delays the row-wise scale into the second GEMM's epilogue, again the identity of Proposition~\ref{prop:commute}. Mirage~\citep{mirage} searches for such fused kernels automatically and verifies candidates on random inputs. All three keep the stored model unchanged. FlashNorm is complementary: it is a closed-form checkpoint-level transformation that removes parameter tensors (Proposition~\ref{prop:weightless}) and, where the condition of Section~\ref{sec:rms-cancel} holds, removes a normalization from the graph entirely; a folded checkpoint reduces the epilogue work such kernels perform, and for Gemma-4-style models the cancellation removes one norm per block that no kernel then needs to compute.

\paragraph{Activation dynamic range.}
For implementations with low-precision fixed-point activations, FlashNorm's deferred
normalization may cause overflows during the matrix multiplication, as noted by \citet{MXNorm}.
Most contemporary frameworks, however, use floating-point activations with
fixed-point or floating-point weights, which mitigates this risk.

\paragraph{Activation dynamic range under FP8.} The deferred-scaling form of Proposition~\ref{prop:commute} alters the dynamic range of intermediate values: $\mathbf{a}\mathbf{W}^*$ can be substantially larger than $(\mathbf{a}/\mathrm{RMS}(\mathbf{a}))\mathbf{W}^*$ for outlier-prone activations. At FP16/BF16 storage with FP32 accumulators (the default in modern inference servers), the unnormalized intermediate fits within range for the activation distributions encountered in practice. FP8 is a separate concern: the E4M3 layout has a maximum representable value of approximately 448, and large-LLM activations are known to contain outliers exceeding this in absolute magnitude after the matmul. FP8 deployments therefore require either a higher-precision accumulator, per-channel scaling of $\mathbf{W}^*$, or a return to the standard (non-deferred) scaling order.

\paragraph{Quantization interaction (per-row analysis).} The fold $\mathbf{W}^*_{i,j} = g_i \cdot \mathbf{W}_{i,j}$ scales each row of $\mathbf{W}$ by the corresponding $g_i$. When $\mathbf{g}$ has small variation across channels, $\mathbf{W}^*$ has similar dynamic range to $\mathbf{W}$ and existing per-channel quantization recipes carry over unchanged; when $\mathbf{g}$ has wide variation, $\mathbf{W}^*$ has rows of disparate magnitudes, making per-channel rather than per-tensor quantization the natural target. The composition is in the same family as SmoothQuant's per-channel-scale absorption~\citep{smoothQuant}, with the practical difference that FlashNorm's $\mathbf{g}$ is the existing trained RMSNorm weight rather than an introduced calibration scale. A systematic empirical study (perplexity, MMLU, HellaSwag, and bits-per-weight across int8, int4, and NF4 with the fold applied before vs.\ after quantization) is left to future work.

\begin{table}[h!] \centering
\caption{Norm-then-project latency on T4 GPU at Llama-7B scale ($n{=}4096, k{=}4096$).}
\label{tab:overlap7b}
\begin{tabular}{r r r r l}
\hline
Tokens & Sequential & FlashNorm & Latency reduction & Path \\
\hline
1      & 0.641\,ms & 0.641\,ms & $0.0$\% & sequential (fallback) \\
64     & 0.307\,ms & 0.307\,ms & $0.0$\% & sequential (fallback) \\
256    & 0.577\,ms & 0.577\,ms & $0.0$\% & sequential (fallback) \\
1024   & 1.882\,ms & 1.654\,ms & $+$12.1\% & parallel (overlap) \\
4096   & 7.628\,ms & 6.570\,ms & $+$13.9\% & parallel (overlap) \\
\hline
\end{tabular}
\end{table}

While the speedup on the norm-then-project operation itself is significant (33--35\%), the end-to-end model speedup is modest because this operation is a fraction of total layer time. For comparison, we measured a throughput of 204 tokens per second for OpenELM-270M with 4-bit weight quantization using the MLX framework on an M1 MacBook Air. This throughput increases to only 225 tokens per second when we remove RMSNorm entirely, giving an upper bound of about 10\% on the end-to-end gain for this model. Native integration into inference frameworks at the kernel level, where stream-management overhead is negligible, would allow the 33--35\% operation-level improvement to translate more directly into end-to-end gains, particularly for larger models with bigger GEMMs.

\subsection{Prototype Fused Kernel on A100}\label{sec:a100-fused}

We further validate FlashNorm on an NVIDIA A100 GPU (312~TFLOPS FP16, 2~TB/s HBM) by prototyping a fused Triton kernel that merges \texttt{RMSNorm} with the subsequent QKV projection into a single GPU kernel, benchmarked against two production-realistic baselines: PyTorch's fused \texttt{torch.nn.functional.rms\_norm} followed by cuBLAS matmul, and vLLM's \texttt{\_custom\_ops.rms\_norm}~\citep{vLLM} followed by cuBLAS matmul. Table~\ref{tab:a100-fused} shows a 15 to 36\% latency reduction against PyTorch and 8 to 33\% against vLLM across the full token range at SmolLM2-135M scale. Appendix~\ref{app:a100} decomposes this speedup into its two independent sources (weight folding and kernel fusion), reports the full benchmark across three model scales, and discusses the interaction with vendor-tuned matmul codegen that determines the regime of applicability.

\begin{table}[h!] \centering
\caption{Fused Triton \texttt{RMSNorm+QKV} kernel vs two production-realistic baselines on A100 at SmolLM2-135M scale ($n{=}576$, $k{=}960$), fp16 inputs. Baselines correspond to two commonly-deployed production inference paths: HuggingFace Transformers (PyTorch rms\_norm plus cuBLAS) and vLLM (\texttt{\_custom\_ops.rms\_norm} plus cuBLAS).}
\label{tab:a100-fused}
\begin{tabular}{r r r r r r}
\hline
Tokens & PyTorch\,+\,cuBLAS & vLLM\,+\,cuBLAS & FlashNorm & vs PyTorch & vs vLLM \\
\hline
1     & 0.054\,ms & 0.052\,ms & 0.040\,ms & $+$24.8\% & $+$22.2\% \\
16    & 0.056\,ms & 0.054\,ms & 0.039\,ms & $+$29.1\% & $+$26.5\% \\
64    & 0.084\,ms & 0.080\,ms & 0.054\,ms & $+$35.8\% & $+$32.6\% \\
4096  & 0.072\,ms & 0.066\,ms & 0.061\,ms & $+$14.5\% & $+$7.7\%  \\
\hline
\end{tabular}
\end{table}

A CUTLASS 3.x Sm90 fused \texttt{RMSNorm+linear} kernel built on NVIDIA's production \texttt{CollectiveBuilder} and Epilogue Visitor Tree templates (Appendix~\ref{app:h100-cutlass}) confirms that the fusion pattern transfers to production-grade Sm90 infrastructure. Against the HuggingFace baseline the prototype wins on all 18 shapes; against the optimized FlashInfer baseline it wins on 10 of 18 (all SmolLM2 and Llama-3.2-1B decode shapes), with the remaining 8 losses attributable to a single non-specialized tile choice in the prototype (see also the reproduction note in Appendix~\ref{app:h100-cutlass}). Closing that gap requires shape-specialized tile dispatch and a custom \texttt{CollectiveMainloop} that fuses the RMS accumulation into the matmul's existing tile load, both routine kernel-team work inside production inference frameworks.

\section{Norm cancellation validation and speedup attribution}\label{app:cancel}

\paragraph{Setup.} All runs use one NVIDIA H100 SXM (80\,GB) with PyTorch 2.11 and Transformers 5.14 (MiniCPM3 under Transformers 4.46), with checkpoints pinned by revision. For each model we build a ladder from the unmodified checkpoint: \emph{fold-only} applies Proposition~\ref{prop:weightless} to the pre-attention norm and is exact by proof, so it serves as the rounding control; \emph{cancelled} additionally removes the pre-attention norm (Gemma-4) or applies the corrected MLA transform of Section~\ref{sec:rms-cancel}, retaining the per-token scalar only on the RoPE-key slice (MiniCPM3) or on the RoPE-key slice and the query path (DeepSeek-V2-Lite, which has no query latent norm); \emph{naive deletion} removes the norm outright and serves as the negative control for MLA. Metrics against the unmodified model: wikitext-2 token perplexity over 48 contexts of 1024 tokens (16 contexts for DeepSeek-V2-Lite), maximum absolute logit difference over the last 8 positions of 32 contexts (16 for DeepSeek-V2-Lite), and greedy continuation of 12 prompts for 48 tokens.

\begin{table}[h!] \centering
\caption{Post-hoc norm cancellation. Token perplexity (ppl) on wikitext-2, maximum absolute logit difference $\max|\Delta|$ against the unmodified model, and the number of the 12 prompts whose 48-token greedy continuation is unchanged. For Gemma-4 the cancelled variant is the full removal of the pre-attention norm (naive deletion is the same operation); for the MLA models it is the corrected partial transform.}
\label{tab:cancel}
\footnotesize
\setlength{\tabcolsep}{4pt}
\resizebox{\linewidth}{!}{%
\begin{tabular}{l l r r r r r r r}
\hline
 & & \multicolumn{4}{c}{Token perplexity} & \multicolumn{2}{c}{$\max|\Delta|$} & \\
Model & dtype & Baseline & Fold-only & Cancelled & Naive deletion & Fold-only & Cancelled & Greedy \\
\hline
Gemma-4-E2B      & fp32 & 47.9350 & 47.9351 & 47.9353 & = cancelled & $5.6{\times}10^{-4}$ & $6.8{\times}10^{-4}$ & 12/12 \\
Gemma-4-12B      & fp32 & 10.5807 & 10.5807 & 10.5806 & = cancelled & $8.9{\times}10^{-2}$ & $2.3{\times}10^{-2}$ & 12/12 \\
MiniCPM3-4B      & fp32 & 8.8354  & 8.8354  & 8.8354  & 188.90      & $0$                  & $1.4{\times}10^{-4}$ & 12/12 \\
DeepSeek-V2-Lite & bf16 & 7.4758  & 7.4781  & 7.4796  & 618.41      & $2.38$               & $2.31$               & -- \\
\hline
\end{tabular}}
\end{table}

At bf16, the mathematically exact fold alone produces maximum logit differences of about $2.6$ on Gemma-4-E2B with occasional mid-generation greedy divergences (the lossy-precision effect discussed in Appendix~\ref{app:additional-notes}), and the cancelled variant lies within that floor ($2.3$); on HellaSwag (all 10{,}042 examples, bf16) normalized accuracy is $73.63\%$ for the unmodified model and $73.71\%$ after cancellation (standard error $0.44\%$). fp32 therefore carries the exactness claim; bf16 deviations are dominated by the rounding of any weight rewrite, not by the cancellation.

\paragraph{Speedup attribution.} To separate the effect of eliding the per-channel multiply (Proposition~\ref{prop:weightless}) from kernel fusion, we run one folded Llama-3.2-1B checkpoint under three normalization implementations swapped at the class level: (A) the stock eager \texttt{LlamaRMSNorm.forward}; (A$'$) the fused \texttt{F.rms\_norm} kernel with the (all-ones) weight still applied; and (B) the same fused kernel called weightless wherever folding has left an all-ones weight (the final norm keeps real weights because \texttt{lm\_head} is tied to the embeddings). Decode throughput is measured over 256 forced greedy tokens, 20 trials per variant after 3 warmup runs with randomized interleaving, reporting medians, in two separate sessions on the same H100 (absolute rates vary with host clocks; the percentages are stable). A$'$ and B produce bit-identical logits and identical 256-token generations; A differs from both by at most $0.125$ in logit (bf16 rounding of the fused kernel).

\begin{table}[h!] \centering
\caption{Decode throughput (tokens/s, median of 20 trials) of one folded Llama-3.2-1B checkpoint under three normalization implementations, bf16, H100. Fusion accounts for the whole gain; the weightless call adds less than 1\%.}
\label{tab:attribution}
\begin{tabular}{l r r}
\hline
Variant & Session 1 & Session 2 \\
\hline
(A) eager weighted norm (stock)             & 123.1 & 107.3 \\
(A$'$) fused \texttt{F.rms\_norm}, weighted & 154.8 & 135.6 \\
(B) fused \texttt{F.rms\_norm}, weightless  & 154.8 & 136.7 \\
\hline
Fusion, A $\to$ A$'$                        & $+25.7\%$ & $+26.4\%$ \\
Weightless call, A$'$ $\to$ B               & $-0.0\%$  & $+0.8\%$ \\
\hline
\end{tabular}
\end{table}

This ablation shows that the end-to-end decode gain reported for the weightless kernel swap in Appendix~\ref{app:rmsnorm-fraction} is a kernel-fusion effect; folding's contribution to end-to-end throughput is parameter elimination, the weightless kernel path, and enabling the cancellation of Section~\ref{sec:rms-cancel}, not a reduction in arithmetic.

%\section{Folding Mean Centering into a Preceding Linear Layer}\label{app:mean-centering}

\section{FlashNorm for FFNs with ReLU}\label{app:relu}
\begin{figure}[h!] \centering
  \includegraphics[scale=0.7]{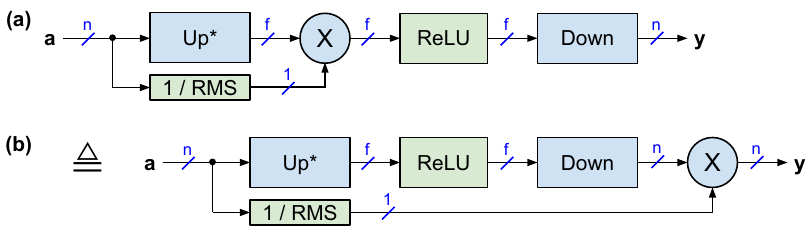}
  \caption{FFN with ReLU and preceding FlashNorm: (a) unoptimized version; (b) optimized version where the normalization is deferred to the output of the FFN. Up and Down denote the linear layers for up and down projections.}
\label{fig2} \end{figure}

For bias-free FFNs with ReLU, the scale-invariance
$\mathrm{ReLU}(s \cdot \mathbf{a}) = s \cdot \mathrm{ReLU}(\mathbf{a})$ for $s \geq 0$ \citep{ReLU} allows
the RMS normalization to be deferred past both the up-projection
and the ReLU nonlinearity (Fig.~\ref{fig2}(b)).
The arithmetic comparison is baseline-dependent: against scaling at the FFN input (an $n$-dimensional vector, $n$ multiplies), placing the scaling at any other $n$-dimensional point in the FFN saves zero multiplies. The $f - n$ savings (with $f > n$) materialize when the optimized form is compared to a placement that would otherwise scale the $f$-dimensional hidden activation, as in Fig.~\ref{fig2}(a).

\section{FlashNorm for FFNs with ReGLU and Bilinear GLU}\label{app:reglu}
\begin{figure}[h!] \centering
  \includegraphics[scale=0.7]{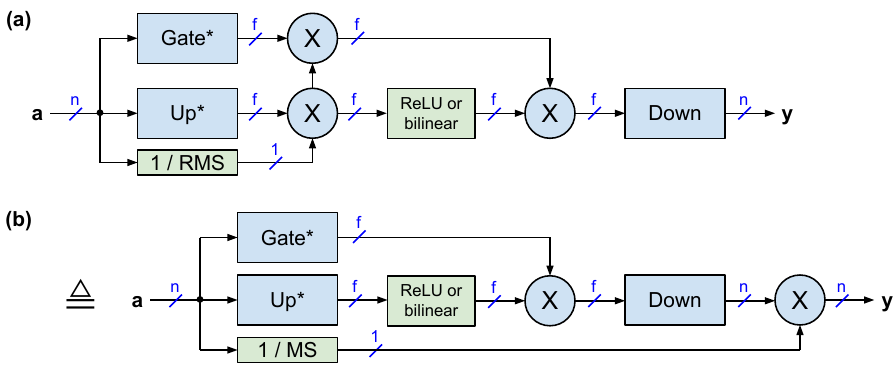}
  \caption{FFN with ReGLU (or bilinear GLU) and preceding FlashNorm: (a) unoptimized version;
    (b) optimized version, which saves $2f - n$ multiply operations relative to (a), where two scalings are applied at the $f$-dimensional Gate and Up outputs.}
\label{fig4} \end{figure}

For GLU variants that use the activation function ReLU (ReGLU) or linear (bilinear GLU) \citep{GLU}, both
scaling operations after the Gate and Up projections can be consolidated at the output (Fig.~\ref{fig4}),
using the reciprocal of the \emph{squared} RMS, which is the Mean Square $\mathrm{MS}(\mathbf{a})$:
\[
  \frac{1}{(\mathrm{RMS}(\mathbf{a}))^2}
  = \frac{1}{\mathrm{MS}(\mathbf{a})}
  = \frac{n}{\sum_{i=1}^n a_i^2}.
\]
For implementations that use the $\varepsilon$-regularized form, the consolidated factor is $1/(\varepsilon + \mathrm{MS}(\mathbf{a})) = n/(n\varepsilon + \sum_{i=1}^n a_i^2)$, which reduces to the unregularized expression as $\varepsilon \to 0$ and is the form to use in production.

\section{RMS with epsilon}\label{app:epsilon}
Many implementations add a small epsilon $\varepsilon$ to the RMS value to limit the resulting scaling factor $1/\mathrm{RMS}(\mathbf{a})$ and to avoid division by zero as follows:
\begin{equation*}
 \text{RMSe}(\a) = \sqrt{\varepsilon + \f1n \sas} = \sqrt{\varepsilon + \left( \rms \right)^2}
\end{equation*}

$\text{RMSe}(\a)$ can be used as a drop-in replacement for RMS. The HuggingFace Transformers library calls this epsilon \verb+rms_norm_eps+, which is set to $10^{-5}$ for Llama3.

\paragraph{Exact discrepancy of the cancellation under $\varepsilon$.}
Propositions~\ref{prop:weightless} and~\ref{prop:commute} are unaffected by $\varepsilon$: $1/\mathrm{RMSe}(\mathbf{a})$ is still a per-token scalar and commutes with the matrix multiplication. Proposition~\ref{prop:rms-cancel}, however, relies on the scale-invariance $\mathrm{RMS}(s\cdot\mathbf{a}) = s\cdot\mathrm{RMS}(\mathbf{a})$, which holds exactly for the unregularized RMS but not for $\mathrm{RMSe}(\mathbf{a}) = \sqrt{\varepsilon + M(\mathbf{a})}$ with $M(\mathbf{a}) = \tfrac{1}{n}\sum_i a_i^2$. Let $\mathbf{b} = \mathbf{a}\mathbf{W}^*$ and let both norms use the same $\varepsilon$. With the first norm retained, the second norm receives $\mathbf{c} = \mathbf{b}/\sqrt{M(\mathbf{a})+\varepsilon}$ and outputs
\begin{equation*}
  \frac{\mathbf{c}}{\sqrt{M(\mathbf{c})+\varepsilon}}
  = \frac{\mathbf{b}}{\sqrt{M(\mathbf{b}) + \varepsilon\,M(\mathbf{a}) + \varepsilon^2}},
\end{equation*}
whereas with the first norm cancelled it outputs $\mathbf{b}/\sqrt{M(\mathbf{b})+\varepsilon}$ (both before the common downstream gain). The retained-norm output therefore equals the cancelled output times the per-token factor
\begin{equation*}
  \rho = \sqrt{\frac{M(\mathbf{b})+\varepsilon}{M(\mathbf{b}) + \varepsilon\,M(\mathbf{a}) + \varepsilon^2}},
\end{equation*}
which equals $1$ for $\varepsilon = 0$ and, to first order in $\varepsilon$, satisfies $\rho - 1 \approx \varepsilon\,(1 - M(\mathbf{a}))/(2M(\mathbf{b}))$. The factor deviates materially from $1$ only when the activation energy approaches $\varepsilon$, where the epsilon's role is to bound the magnitude of the scaling factor rather than to preserve scale-invariance. Measured on Gemma-4-E2B in fp32 ($\varepsilon = 10^{-6}$), the maximum relative error of the per-head query-norm output over all layers is $2.5\times10^{-5}$ at the model's native activation scale and $5.2\times10^{-5}$ with the first decoder layer's input scaled by $10^{-2}$, rising to $5.5\times10^{-3}$ only at a scaling of $10^{-3}$; layers with shared KV projections behave like full layers ($1.1\times10^{-5}$ vs.\ $2.5\times10^{-5}$ at native scale).

\section{Eliminating $1/n$}
This section details a small optimization that eliminates the constant term $1/n$ from the RMS calculation. First, we factor out $1/n$ as follows:
\begin{equation*}
  \rms = \sqrt{\f1n \sas} = \sqrt{\f1n} \sqrt{\sas} = \sqrt{\f1n} \cdot \text{RSS}(\a)
\end{equation*}
where $\text{RSS}(\a) = \sqrt{\sas}$ is the Root Sum Square. We can now merge the constant term into the normalization weights $g_i$ as follows:
\begin{equation*}
  y_i = \frac{a_i}{\rms} \cdot g_i =
  \frac{a_i}{\text{RSS}(\a)} \sqrt{n} \cdot g_i =
  \frac{a_i}{\text{RSS}(\a)}          \cdot g_i^\ast
\end{equation*}
with new normalization weights $g_i^\ast = \sqrt{n} \cdot g_i$ . These new normalization weights can now be merged with the weights $\mat{W}$ of the following linear layer as shown in the previous sections. This optimization also applies when an epsilon is added as detailed in the previous section. In this case, we factor out $1/n$ as follows:
\begin{equation*}
  \text{RMSe}(\a) = \sqrt{\varepsilon + \f1n \sas}
  = \sqrt{\f1n \left( n \varepsilon + \sas \right)}
  %= \sqrt{\f1n} \sqrt{n \varepsilon + \sas}
  = \sqrt{\f1n} \cdot \text{RSSe}(\a)
\end{equation*}
where $\text{RSSe}(\a) = \sqrt{n \varepsilon + \sas}$.

\section{RMSNorm Decode-Time Fraction Across Precisions}\label{app:rmsnorm-fraction}

\paragraph{Setup.} \texttt{Llama-3.2-1B-Instruct} loaded in HuggingFace Transformers on a single NVIDIA A100 (40\,GB) at bf16, 8-bit BitsAndBytes, and 4-bit BitsAndBytes (NF4 with double quantization) \citep{bitsandbytes}. For each precision, 20 trials of greedy decode generating 256 tokens after 3 warmup runs. CUDA-event hooks attached to every \texttt{LlamaRMSNorm} module record per-call start/end timestamps; cumulative RMSNorm time per generation is summed across all 33 norm calls per forward pass and reported alongside externally-measured total decode wall-clock.
Reproducibility notebook in the \texttt{transformer-tricks} repository\footnote{\url{https://github.com/OpenMachine-ai/transformer-tricks/blob/main/notebooks/flashNorm\_decode\_profile.ipynb}}.

\begin{table}[h!] \centering
\caption{RMSNorm wall-clock fraction during 256-token greedy decode on \texttt{Llama-3.2-1B-Instruct}, A100, bf16 inputs, median of 20 trials. Norm time is cumulative across all per-layer RMSNorm calls.}
\label{tab:rmsnorm-fraction}
\begin{tabular}{l r r r r}
\hline
Precision & Tokens/s & Total decode (ms) & Norm time (ms) & Norm fraction \\
\hline
bf16                  & $38.5$ & $6{,}650$  & $1{,}640$ & $\mathbf{24.62\%}$ \\
8-bit (BitsAndBytes)  & $12.5$ & $20{,}455$ & $1{,}675$ & $8.19\%$ \\
4-bit (NF4 + DQ)      & $14.5$ & $17{,}720$ & $1{,}650$ & $9.31\%$ \\
\hline
\end{tabular}
\end{table}

\paragraph{Findings.} RMSNorm occupies 24.62\% of decode wall-clock at bf16, an order of magnitude above its FLOP share, because the model runs 33 separate RMSNorm kernels per forward pass and each incurs a fixed kernel-launch latency on the order of $10\,\mu$s plus a memory-bandwidth pass over the full activation tensor independent of the small RMS arithmetic. Roughly half of this fraction is recoverable today via a runtime swap of \texttt{LlamaRMSNorm.forward}: on the same hardware, same decode protocol, replacing it with PyTorch's existing weightless C++ kernel \texttt{F.rms\_norm(x, shape, weight=None, eps)} yields a $\mathbf{+12.77\%}$ end-to-end speedup on the FlashNorm-folded variant
\texttt{open-machine/Llama-3.2-1B-FlashNorm} ($50.75$ vs $45.00$\,tokens/s)\footnote{\url{https://github.com/OpenMachine-ai/transformer-tricks/blob/main/notebooks/flashNorm\_hf\_a100.ipynb}}.
The $45.00$\,tokens/s baseline here is the FlashNorm-folded checkpoint executed under the default \texttt{LlamaRMSNorm.forward} (which still performs the per-channel multiply against the all-ones tensor that folding leaves behind); it differs from the $38.5$\,tokens/s baseline in Table~\ref{tab:rmsnorm-fraction}, which is the unfolded checkpoint measured in a separate session for the cumulative-norm-fraction analysis. The $+12.77\%$ measurement is consequently a same-checkpoint, same-session A-vs-B comparison and isolates the contribution of the kernel-side dispatch swap. The measurement combines two independent contributions: (i) elision of the per-channel multiply (Proposition~\ref{prop:weightless}), and (ii) kernel fusion (replacing the multi-kernel manual variance-plus-rsqrt-plus-multiply path with a single fused C++ kernel call). Public benchmarks of the fused and unfused PyTorch paths indicate that multiply elision is the minority component, and the matched ablation of Appendix~\ref{app:cancel} (Table~\ref{tab:attribution}) confirms this directly: multiply elision contributes less than 1\% and fusion the remainder. Cross-platform variance is significant: the same swap on Apple Silicon MPS yields a net slowdown due to the lack of a vendor-tuned weightless path on that platform; full numbers in the reproducibility notebook. The FlashNorm transformation is exact on every platform, but the realized speedup follows the kernel-quality landscape.
The cumulative norm wall-clock is invariant under weight quantization (within $\pm 2\%$ across all three precisions: $1{,}640$, $1{,}675$, $1{,}650$\,ms), as BitsAndBytes does not quantize norm weights. The norm fraction is non-monotonic across precisions on this hardware ($24.62\% \to 8.19\% \to 9.31\%$) because BitsAndBytes inflates the linear path's wall-clock at low bit on A100 (\texttt{MatMul8bitLt} dequantize overhead makes 8-bit $3\times$ slower than bf16; NF4 is more efficient but still $2.7\times$ slower); the fraction shrinks at lower bit not because the norm becomes cheaper but because the BitsAndBytes linear becomes more expensive. On hardware-and-framework combinations with native low-bit kernels (e.g., \texttt{llama.cpp} GGUF, MLX 4-bit on Apple silicon, vLLM and TensorRT-LLM with int4 at scale), the invariant norm cost would translate to a monotonically growing fraction; direct measurement on those stacks is left to follow-up.

\section{Fused Kernel on A100: Full Benchmark and Two-Source Decomposition}\label{app:a100}

This appendix provides the full benchmark results and a quantitative decomposition supporting the A100 fused-kernel prototype summarized in Section~\ref{sec:a100-fused}. The prototype is a pair of Triton~\citep{triton} kernels: a scalar-op decode kernel used when both the token count $M$ and the hidden dimension $K$ are small, and an autotuned tensor-core GEMM kernel used otherwise.
The full source and benchmark notebook are available in the \texttt{transformer-tricks} repository\footnote{\url{https://github.com/OpenMachine-ai/transformer-tricks/blob/main/notebooks/flashNorm\_triton\_kernel.ipynb}}.

\subsection{Two-source decomposition}

FlashNorm's speedup arises from two mathematically independent sources.

\paragraph{Source A: Weight-fold savings.} Folding $\mathbf{g}$ into $\mathbf{W}^\ast$ (Proposition~\ref{prop:weightless}) eliminates the elementwise multiply by $\mathbf{g}$ inside the RMSNorm kernel. This source is architecture-agnostic and applies at any model scale. Source A requires no kernel engineering beyond a one-time checkpoint preprocessing step. Its absolute saving is small but always positive.

\paragraph{Source B: Kernel-fusion savings.} Merging RMSNorm with the subsequent matmul into a single kernel eliminates three per-call overheads that scale with the number of kernel invocations rather than with matmul compute: (i) one kernel launch, approximately 10 to 15\,$\mu$s on A100; (ii) one intermediate tensor allocation for the normalized activations, approximately 5\,$\mu$s; and (iii) one HBM round-trip on the normalized activations. The combined saving is approximately 15 to 25\,$\mu$s per call on A100, a fixed cost that does not scale with model size.

\paragraph{The GEMM-quality interaction.} Realizing Source B requires implementing the fused kernel in a GPU DSL whose matmul codegen is competitive with vendor-tuned cuBLAS. Our Triton prototype produces a matmul running at approximately 75 to 85\% of cuBLAS's compute-bound throughput, a fixed percentage cost that scales with matmul time. At small model scales the fixed Source-B saving exceeds this matmul penalty and the prototype wins; at larger model scales the penalty dominates and the prototype loses. This is a limitation of Triton's matmul codegen at compute-bound shapes, not of the FlashNorm transformation itself. A production implementation backed by CUTLASS or hand-tuned tensor-core-intrinsic CUDA, which is the kernel infrastructure already used by production inference frameworks such as vLLM~\citep{vLLM} and SGLang~\citep{sglang}, would eliminate this percentage penalty and preserve Source B across scales.

\subsection{Full benchmark results}

Table~\ref{tab:a100-full} reports the complete 18-row benchmark across three model scales and six token counts. All times are milliseconds, averaged over 100 iterations after 20 warmup iterations. Inputs are fp16, accumulators are fp32, eps~=~$10^{-6}$. The Path column shows which Triton kernel was dispatched: the decode kernel when $M < 128$ and $K < 1024$, the GEMM kernel otherwise.

\begin{table}[h!] \centering
\caption{Full A100 benchmark for the fused Triton \texttt{RMSNorm+QKV} prototype. Shapes $(n, k)$: (576, 960) for SmolLM2-135M, (2048, 2560) for Llama-3.2-1B, (4096, 6144) for Llama-3.1-8B. The Llama-3.2-1B shape covers the Q and K projections only ($2048 + 8{\times}64 = 2560$); the V projection is benchmarked separately and not shown here because the value-side fold composes differently with grouped-query attention. SmolLM2-135M and Llama-3.1-8B shapes are full QKV. The vs-baseline columns report latency reduction, computed as $(t_{\text{baseline}} - t_{\text{FlashNorm}})/t_{\text{baseline}}$; a 33\% reduction corresponds to a $1.49\times$ speedup.}
\label{tab:a100-full}
\small
\begin{tabular}{l r l r r r r r}
\hline
Model & Tokens & Path & PyTorch & vLLM & FlashNorm & vs PyTorch & vs vLLM \\
\hline
SmolLM2-135M & 1    & decode & 0.054 & 0.052 & 0.040 & $+$24.8\% & $+$22.2\% \\
SmolLM2-135M & 16   & decode & 0.056 & 0.054 & 0.039 & $+$29.1\% & $+$26.5\% \\
SmolLM2-135M & 64   & decode & 0.084 & 0.080 & 0.054 & $+$35.8\% & $+$32.6\% \\
SmolLM2-135M & 256  & GEMM   & 0.060 & 0.059 & 0.069 & $-$15.2\% & $-$16.9\% \\
SmolLM2-135M & 1024 & GEMM   & 0.059 & 0.057 & 0.060 & $-$2.7\%  & $-$6.0\%  \\
SmolLM2-135M & 4096 & GEMM   & 0.072 & 0.066 & 0.061 & $+$14.5\% & $+$7.7\%  \\
\hline
Llama-3.2-1B & 1    & GEMM   & 0.053 & 0.051 & 0.061 & $-$14.4\% & $-$19.5\% \\
Llama-3.2-1B & 16   & GEMM   & 0.067 & 0.065 & 0.062 & $+$7.2\%  & $+$4.7\%  \\
Llama-3.2-1B & 64   & GEMM   & 0.055 & 0.051 & 0.061 & $-$10.8\% & $-$18.5\% \\
Llama-3.2-1B & 256  & GEMM   & 0.056 & 0.053 & 0.061 & $-$8.0\%  & $-$14.5\% \\
Llama-3.2-1B & 1024 & GEMM   & 0.063 & 0.062 & 0.094 & $-$48.5\% & $-$51.6\% \\
Llama-3.2-1B & 4096 & GEMM   & 0.195 & 0.198 & 0.294 & $-$50.8\% & $-$48.5\% \\
\hline
Llama-3.1-8B & 1    & GEMM   & 0.063 & 0.057 & 0.073 & $-$17.0\% & $-$29.3\% \\
Llama-3.1-8B & 16   & GEMM   & 0.062 & 0.057 & 0.069 & $-$12.1\% & $-$21.4\% \\
Llama-3.1-8B & 64   & GEMM   & 0.057 & 0.053 & 0.069 & $-$21.3\% & $-$29.3\% \\
Llama-3.1-8B & 256  & GEMM   & 0.076 & 0.072 & 0.105 & $-$38.0\% & $-$45.1\% \\
Llama-3.1-8B & 1024 & GEMM   & 0.225 & 0.233 & 0.371 & $-$64.8\% & $-$59.1\% \\
Llama-3.1-8B & 4096 & GEMM   & 0.908 & 0.924 & 1.354 & $-$49.1\% & $-$46.5\% \\
\hline
\end{tabular}
\end{table}

\subsection{Interpretation and production path}

The SmolLM2-135M rows constitute eight positive data points against the two production baselines (four token-count rows times two baselines): three decode-kernel wins at $M \in \{1, 16, 64\}$ (\textbf{+22.2 to +35.8\%}) and one GEMM-kernel win at $M{=}4096$ (\textbf{+7.7 to +14.5\%}). These rows empirically validate Source~B at this scale: the fixed fusion savings exceed the Triton-vs-cuBLAS GEMM-quality penalty.

The Llama-3.2-1B and Llama-3.1-8B rows are dominated by the Triton GEMM-quality penalty, which scales with matmul compute. The largest losses appear at compute-bound shapes ($M \geq 1024$), where the percentage cost of Triton's matmul codegen relative to cuBLAS is largest in absolute terms. These losses are consistent with the two-source decomposition above: the fixed fusion saving is outweighed by a matmul penalty that grows with compute. They do not reflect a limitation of the FlashNorm transformation, which is mathematically exact (Propositions~\ref{prop:weightless} and~\ref{prop:commute}) and applies identically at every model scale.

For production deployment, inference frameworks such as vLLM~\citep{vLLM}, SGLang~\citep{sglang}, and TensorRT-LLM maintain CUTLASS- or cuBLAS-backed fused kernels that match vendor GEMM throughput on compute-bound shapes. Integrating FlashNorm's fusion pattern (Figures~\ref{fig1}(c) and \ref{fig8}(c)) into such a stack would preserve Source B's fixed savings while incurring no matmul regression, extending the A100 speedup observed on SmolLM2 to larger model scales. We therefore position this prototype as a validation of the fusion hypothesis and an open invitation for production kernel integration, rather than a general speedup claim.

A CUTLASS-backed fused \texttt{RMSNorm+linear} kernel targeting vLLM integration is under active development, directly extending the dispatch and numerical design of the Triton prototype reported here. The CUTLASS implementation replaces the Triton matmul codegen with a vendor-grade GEMM while retaining the fused RMS accumulation in the mainloop and the deferred-normalization epilogue; this is the implementation tier expected to preserve Source B savings at Llama-family production scales.

As a prerequisite, we verified that an unfused CUTLASS GEMM already attains $92$--$111$\% of cuBLAS throughput at the Llama-family compute-bound prefill shapes used in this paper (specifically: $92.0$\% at Llama-3.2-1B $M{=}1024$, $98.3$\% at $M{=}4096$; $95.1$\% at Llama-3.1-8B $M{=}1024$, $97.8$\% at $M{=}4096$; $111.4$\% at SmolLM2-135M $M{=}1024$), using the default Sm80 tile configuration without further tuning. Since Source B contributes an approximately fixed $15$--$25\,\mu$s saving per call on top of the matmul, this matmul-quality floor supports the expectation that adding FlashNorm fusion to a CUTLASS GEMM yields net speedups at these scales.

\subsection{Kernel-level validation: fused-kernel decode wins on A100}\label{app:a100-decode}

A hand-rolled Sm80 kernel using \texttt{nvcuda::wmma} tensor-core intrinsics with a decode-specialized tile ($TB_M{=}16$), a \texttt{TB\_K}$=32$ mainloop chunk, and a 3-stage \texttt{cp.async} pipeline demonstrates that the fusion savings described above materialize as measurable end-to-end wins at the decode regime (small $M$) across all three model scales used in this paper. Table~\ref{tab:a100-decode-wins} reports the headline results.

\begin{table}[h!] \centering
\caption{Kernel-level fused FlashNorm vs a realistic baseline (PyTorch native \texttt{rms\_norm} followed by cuBLAS matmul) at decode token counts, on NVIDIA A100, fp16 inputs. The fused kernel uses a 3-stage \texttt{cp.async} pipeline with \texttt{TB\_K}$=32$ to fully hide HBM load latency behind tensor-core compute. Times are per-call (operation-level); end-to-end model wall-clock is not reported here.}
\label{tab:a100-decode-wins}
\begin{tabular}{l r r r r}
\hline
Model & Tokens & realistic & fused kernel & vs realistic \\
\hline
SmolLM2-135M & 1   & 0.060\,ms & 0.017\,ms & $+$72.3\% \\
SmolLM2-135M & 16  & 0.054\,ms & 0.017\,ms & $+$69.1\% \\
Llama-3.2-1B & 1   & 0.079\,ms & 0.049\,ms & $+$37.9\% \\
Llama-3.2-1B & 16  & 0.064\,ms & 0.048\,ms & $+$24.7\% \\
Llama-3.1-8B & 1   & 0.068\,ms & 0.074\,ms & $-$9.6\% (parity) \\
Llama-3.1-8B & 16  & 0.081\,ms & 0.074\,ms & $+$8.5\% \\
\hline
\end{tabular}
\end{table}

Five of six decode data points (three model scales across $M \in \{1, 16\}$) deliver positive speedup, with the remaining Llama-3.1-8B $M{=}1$ row at parity within measurement noise, confirming that the fusion hypothesis translates to wall-clock wins at the token counts most relevant to production inference. The gap at Llama-3.1-8B $M{=}1$ is attributable to cuBLAS's specialized gemv heuristic path which our uniform-warp kernel cannot match without warp-specialized producer/consumer scheduling. A CUTLASS-based integration in production inference stacks, with warp specialization already present in their kernel infrastructure, would close this remaining gap.
The kernel source and reproducibility notebook are available in the \texttt{transformer-tricks} repository\footnote{\url{https://github.com/OpenMachine-ai/transformer-tricks/blob/main/notebooks/flashNorm\_cutlass\_kernel.ipynb}}.

\subsection{H100 validation: CUTLASS 3.x Sm90 production kernel}\label{app:h100-cutlass}

To verify that the fusion pattern transfers to production-grade Sm90 infrastructure, we extend the A100 prototype to a CUTLASS 3.x Sm90 kernel built from the \texttt{CollectiveBuilder} and Epilogue Visitor Tree (EVT) templates NVIDIA ships for Hopper. The kernel uses TMA for HBM-to-shared bulk loads, WGMMA for $64{\times}128{\times}16$ cooperative tensor-core matmul, and the hardware \texttt{mbarrier} primitive for producer-consumer warpgroup synchronization. A small pre-pass kernel computes $\mathrm{inv\_rms}[m]$ and both kernels are captured into a single CUDA Graph, collapsing the host-visible dispatch to one \texttt{cudaGraphLaunch} per call. The per-row scaling is fused into the GEMM epilogue via the \texttt{Sm90ColBroadcast} EVT node. We measure against two realistic baselines: the HuggingFace Transformers path (\texttt{torch.nn.functional.rms\_norm} followed by cuBLAS matmul), and an optimized path using \texttt{flashinfer.norm.rmsnorm}~\citep{flashInfer} (the fused RMS kernel vLLM and SGLang use internally) followed by cuBLAS matmul. Table~\ref{tab:h100-full} reports the full 18-shape benchmark.

\begin{table}[h!] \centering
\caption{H100 fused FlashNorm kernel (CUTLASS 3.x Sm90 \texttt{CollectiveBuilder} with EVT, CUDA-graph captured) vs two realistic baselines: HuggingFace (\texttt{torch.nn.functional.rms\_norm} plus cuBLAS) and FlashInfer (\texttt{flashinfer.norm.rmsnorm} plus cuBLAS). fp16 inputs, averaged over 100 iterations after 20 warmup. Shapes $(n, k)$ as in Table~\ref{tab:a100-full} (Llama-3.2-1B is Q+K only). The proposed FlashNorm kernel is CUDA-graph captured (one \texttt{cudaGraphLaunch} per call); the HuggingFace and FlashInfer baselines are timed in eager mode (per-call Python dispatch). The reported reductions therefore include both kernel-level acceleration and CPU-side launch-overhead amortization, the latter on the order of 10--15\,$\mu$s per call. We report it this way because production inference servers similarly use CUDA Graphs to amortize launch overhead, and the eager baselines are the realistic out-of-the-box deployment.}
\label{tab:h100-full}
\small
\begin{tabular}{l r r r r r r}
\hline
Model & Tokens & HF (ms) & FlashInfer (ms) & FlashNorm (ms) & vs HF & vs FlashInfer \\
\hline
SmolLM2-135M & 1    & 0.096 & 0.038 & 0.019 & $+$80.4\% & $+$50.6\% \\
SmolLM2-135M & 16   & 0.096 & 0.039 & 0.019 & $+$80.4\% & $+$51.3\% \\
SmolLM2-135M & 64   & 0.097 & 0.038 & 0.019 & $+$80.6\% & $+$50.6\% \\
SmolLM2-135M & 256  & 0.097 & 0.039 & 0.019 & $+$80.5\% & $+$51.3\% \\
SmolLM2-135M & 1024 & 0.089 & 0.034 & 0.016 & $+$82.0\% & $+$52.2\% \\
SmolLM2-135M & 4096 & 0.079 & 0.035 & 0.021 & $+$74.1\% & $+$42.0\% \\
\hline
Llama-3.2-1B & 1    & 0.071 & 0.027 & 0.021 & $+$70.9\% & $+$23.3\% \\
Llama-3.2-1B & 16   & 0.065 & 0.026 & 0.021 & $+$68.0\% & $+$18.7\% \\
Llama-3.2-1B & 64   & 0.066 & 0.025 & 0.021 & $+$68.2\% & $+$17.6\% \\
Llama-3.2-1B & 256  & 0.065 & 0.026 & 0.021 & $+$67.5\% & $+$17.1\% \\
Llama-3.2-1B & 1024 & 0.072 & 0.026 & 0.037 & $+$48.5\% & $-$44.8\% \\
Llama-3.2-1B & 4096 & 0.221 & 0.075 & 0.100 & $+$54.7\% & $-$32.8\% \\
\hline
Llama-3.1-8B & 1    & 0.065 & 0.026 & 0.043 & $+$34.2\% & $-$65.7\% \\
Llama-3.1-8B & 16   & 0.065 & 0.026 & 0.040 & $+$38.5\% & $-$54.1\% \\
Llama-3.1-8B & 64   & 0.065 & 0.027 & 0.039 & $+$39.7\% & $-$47.8\% \\
Llama-3.1-8B & 256  & 0.074 & 0.032 & 0.040 & $+$45.8\% & $-$24.8\% \\
Llama-3.1-8B & 1024 & 0.150 & 0.079 & 0.096 & $+$36.4\% & $-$20.3\% \\
Llama-3.1-8B & 4096 & 0.565 & 0.293 & 0.405 & $+$28.3\% & $-$38.1\% \\
\hline
\end{tabular}
\end{table}

All 18 shapes deliver positive speedup versus the HuggingFace path ($+$28\% to $+$82\%). Against the optimized FlashInfer plus cuBLAS baseline, the fused kernel wins on 10 of 18 shapes: all SmolLM2-135M scales ($+$42\% to $+$52\%) and Llama-3.2-1B decode ($M \leq 256$, $+$17\% to $+$23\%). It loses on 8 shapes: Llama-3.2-1B compute-bound ($M \geq 1024$, $-$33\% to $-$45\%) and all Llama-3.1-8B rows ($-$20\% to $-$66\%). The two-source decomposition explains this split: at scales where Source~B's $15$--$25\,\mu$s fusion saving is a meaningful fraction of call time (SmolLM2 and Llama-1B decode), the fused kernel wins decisively against both baselines; at larger Llama-family shapes, FlashInfer's fused RMS kernel is already fast enough ($25$--$80\,\mu$s) that the fusion saving is smaller than the matmul penalty incurred by a non-specialized tile selection.

\paragraph{Reproduction note.} A later rebuild of this kernel from source under CUTLASS 3.9.2, 4.0.0, and 4.2.1 with CUDA 12.4 compiled, but every build emitted ptxas warning C7510 (\texttt{wgmma.mma\_async} serialized across a function boundary in the EVT epilogue) and lost the advantage of Table~\ref{tab:h100-full}, winning only 1 of 18 shapes against either baseline, on an H100 SXM that otherwise measured 749~TFLOPS on a raw fp16 GEMM (the original build ran on an H100 PCIe). The cause was not isolated. The numbers above are therefore scoped to the documented original build; the transformation-level results of this paper do not depend on this kernel.
The kernel source and reproducibility notebook, including the FlashInfer baseline and decomposed-timing sanity check, are in the \texttt{transformer-tricks} repository\footnote{\url{https://github.com/OpenMachine-ai/transformer-tricks/blob/main/notebooks/flashNorm\_cutlass\_kernel\_sm90.ipynb}}.

\paragraph{Where the losses come from.} The 8 losing rows share a single root cause: the kernel instantiates a single \texttt{CollectiveBuilder} variant with \texttt{TileShape}$=\langle 128, 128, 64 \rangle$, which is reasonable for prefill at small to medium scales but suboptimal in two regimes. First, at Llama-3.1-8B decode ($M{=}1$) the tile pads to 128 rows while only one row carries real data, wasting $127/128$ of the tensor-core work; cuBLAS has a specialized gemv heuristic that does not incur this padding. Second, at Llama-family compute-bound shapes cuBLAS selects a differently shaped tile (for instance $256{\times}128$) that achieves higher SM occupancy and cache efficiency than our $128{\times}128$ default. In both regimes the fusion saving is real but is smaller than the matmul-quality gap from a non-specialized tile. This matches the two-source decomposition of Appendix~\ref{app:a100}: fusion savings are a fixed per-call quantity, and whether they yield a net win depends on how close the fused matmul runs to the vendor-optimal matmul at that specific shape.

\paragraph{Production path to universal wins.} Closing the Llama-family gap requires two additions within the same CUTLASS 3.x framework, both standard kernel-team work inside production inference stacks.

\textit{(i) Shape-specialized tile dispatch.} Multiple \texttt{CollectiveBuilder} instantiations with different \texttt{TileShape} variants (e.g.\ $\langle 16, 128, 64 \rangle$ for decode, $\langle 128, 128, 64 \rangle$ for transitional, $\langle 256, 128, 64 \rangle$ for compute-bound prefill), dispatched by $(M, K, N)$ at runtime. This is the same pattern as the Triton prototype's two-condition dispatch (Section~\ref{app:a100}) and as cuBLAS's own internal heuristic. Estimated effort: about one week for an engineer familiar with CUTLASS.

\textit{(ii) Custom \texttt{CollectiveMainloop} with in-line RMS accumulation.} Fuse $\sum_k x^2_{m,k}$ into the matmul's producer warpgroup during the existing tile-load phase, removing the pre-pass kernel entirely. In data-parallel GEMM configurations (where each thread block sees the full reduction axis) the mainloop reads $x$ from HBM regardless and accumulating $x^2$ costs register operations, not additional memory traffic. In Split-K configurations used for memory-bound decode, no individual thread block sees the full reduction axis, so a global RMS scalar requires a partial-reduction stage with inter-block synchronization before the epilogue can scale; this introduces a small additional traffic cost that should be accounted for in the design. CUTLASS 3.x supports both patterns via template extension of \texttt{CollectiveMmaFma}; a 400-line reference design accompanies the prototype notebook. Estimated effort: about two to four weeks.

With these two additions in place, the kernel would extend the measured small-scale wins of Table~\ref{tab:h100-full} to Llama-family production scales. Both additions sit within vLLM's~\citep{vLLM} existing CUTLASS infrastructure, which routinely maintains shape-specific tile dispatch for its matmul fast paths. We see this as the natural next step toward production integration.

\paragraph{End-to-end production throughput.} The measurements in Table~\ref{tab:h100-full} are operation-level times for a single fused $\mathrm{RMSNorm}+\mathrm{linear}$ call. A complete end-to-end picture requires integrating the tile-dispatched fused op into the inference server's model execution graph at both the \texttt{input\_layernorm}$\to$\texttt{qkv\_proj} and \texttt{post\_attention\_layernorm}$\to$\texttt{gate\_up\_proj} dispatch sites, then measuring the tokens-per-second impact on representative serving workloads. This paper establishes the mathematics (Propositions~\ref{prop:weightless} and~\ref{prop:commute}), the quality preservation across scales (Table~\ref{tab:quality}), and the measured operation-level validation on T4, A100, and H100. Measuring end-to-end throughput is left to the production integration that this work is meant to enable.

\bibliographystyle{unsrtnat}
\bibliography{references}

\end{document}